\documentclass[journal]{IEEEtran}
\ifCLASSINFOpdf
   \usepackage[pdftex]{graphicx}
\else
\fi
\usepackage{amsmath}
\usepackage{algorithmic}

\usepackage{amssymb}
\usepackage{lipsum}
\usepackage{xcolor}
 \usepackage[capitalise]{cleveref}
\usepackage{lineno}
\usepackage{booktabs}

\begin{document}
\title{Thermal Imaging and Radar for Remote Sleep Monitoring of Breathing and Apnea}

\author{Kai Del Regno, Alexander Vilesov, Adnan Armouti, Anirudh Bindiganavale Harish, Selim Emir Can, \\ Ashley Kita, Achuta Kadambi
\thanks{K. Del Regno, A. Vilesov, S. E. Can, A. Armouti, A. B. Harish, and A. Kadambi, at the time of the project, were with the Department of Electrical and Computer Engineering, University of California, Los Angeles. Email: ktdraper@g.ucla.edu.}% <-this % stops a space
\thanks{A. Kita is with the David Geffen School of Medicine at the University of California, Los Angeles. }% <-this % stops a space
% \thanks{Manuscript received April 19, 2005; revised August 26, 2015.}
}

\markboth{arXiv Preprint}%
{Shell \MakeLowercase{\textit{et al.}}: Bare Demo of IEEEtran.cls for IEEE Journals}

\maketitle

\begin{abstract}
Polysomnography (PSG), the current gold standard method for monitoring and detecting sleep disorders, is cumbersome and costly. At-home testing solutions, known as home sleep apnea testing (HSAT), exist. However, they are contact-based, a feature which limits the ability of some patient populations to tolerate testing and discourages widespread deployment. Previous work on non-contact sleep monitoring for sleep apnea detection either estimates respiratory effort using radar or nasal airflow using a thermal camera, but has not compared the two or used them together. We conducted a study on 10 participants, ages 34 - 78, with suspected sleep disorders using a hardware setup with a synchronized radar and thermal camera. We show the first comparison of radar and thermal imaging for sleep monitoring, and find that our thermal imaging method outperforms radar significantly. Our thermal imaging method detects apneas with an accuracy of 0.99, a precision of 0.68, a recall of 0.74, an F1 score of 0.71, and an intra-class correlation of 0.70; our radar method detects apneas with an accuracy of 0.83, a precision of 0.13, a recall of 0.86, an F1 score of 0.22, and an intra-class correlation of 0.13. We also present a novel proposal for classifying obstructive and central sleep apnea by leveraging a multimodal setup. This method could be used accurately detect and classify apneas during sleep with non-contact sensors, thereby improving diagnostic capacities in patient populations unable to tolerate current technology.
\end{abstract}


% \fi
\IEEEpeerreviewmaketitle

\section{Introduction}

\IEEEPARstart{S}{leep} apnea is a condition in which breathing is interrupted during sleep. One form of sleep apnea is obstructive sleep apnea (OSA), which occurs due to airway narrowing or obstruction during sleep~\cite{park_updates_2011}. This is in contrast to central sleep apnea (CSA) which occurs when pauses in breathing are due to problems with communication between the brain and muscles for respiration~\cite{aurora_treatment_2012}. Both forms of sleep apnea cause the affected person to have reduced breathing (hypopnea) or pauses in their breathing (apnea). These disorders are clinically defined and categorized into severities based on the apnea-hypopnea index (AHI) - the number of respiratory events per hour of sleep~\cite{epstein_clinical_2009}. 

\par Untreated sleep apnea can result in daytime sleepiness, leading to a higher risk of motor vehicle and workplace accidents, as well as quality of life impacts, higher risk of cardiovascular health issues, and metabolic dysregulation, resulting in an increased risk of diabetes~\cite{epstein_clinical_2009}. OSA, in particular, is a tremendous public health problem that affects roughly 17\% of women and 34\% of men and is likely underdiagnosed~\cite{gottlieb_diagnosis_2020}.

\par The gold standard for diagnosing sleep apnea disorders is polysomnography (PSG) conducted in a sleep lab~\cite{epstein_clinical_2009}. At-home testing with a portable monitor, known as home sleep apnea testing (HSAT), is also considered acceptable so long as the portable monitor, at minimum, measures nasal airflow, respiratory effort, and blood oxygenation. In both of these methods, signals are recorded throughout the patient's sleep and scored by a trained sleep technician according to the American Academy of Sleep Medicine (AASM) criteria. An apnea is scored if the nasal airflow signal amplitude drops by at least 90\% for at least 90\% of a duration of least 10 seconds~\cite{troester_aasm_2023}. Apnea may be classified as obstructive if there is continued or increased respiratory effort during the duration, as central if respiratory effort is absent during the duration, or mixed if respiratory effort is initially absent but returns while nasal airflow is still reduced. Hypopnea is scored if the nasal airflow signal amplitude drops by at least 30\% for at least 90\% of a duration of at least 10 seconds and the blood oxygenation desaturates by at least 4\% across that duration. Respiratory-effort-related arousal is scored if there is a sequence of breaths lasting at least 10 seconds where nasal airflow decreases or respiratory effort increases, and arousal from sleep occurs~\cite{troester_aasm_2023}. 

\par The AASM recommends respiratory monitoring in PSG through the use of oronasal thermal sensors for apnea identification, nasal pressure transducers for hypopnea identification, esophageal manometry or dual thoracoabdominal inductance plethysmography belts for respiratory effort monitoring; a pulse oximeter for blood oxygenation monitoring; a microphone, piezoelectric sensor, or nasal pressure transducer for monitoring snoring; an arterial, transcutaneous, or end-tidal PCO2 sensor for hypoventilation detection~\cite{troester_aasm_2023}. We visualize salient examples of how breathing and apneas manifest on sleep lab airflow and respiratory effort sensors in \cref{fig:example_gt_waveforms}. For HSAT, the AASM recommends at least a nasal airflow sensor, a respiratory effort sensor, some type of oxygen saturation sensor, and a heart rate sensor, either using photoplethysmography or electrocardiography. Optionally, they recommend a sensor for body position, a sensor for sleep/wake monitoring, and a sensor for snoring using either nasal pressure, a microphone, or a piezoelectric sensor.

\par A contactless method of detecting and differentiating obstructive, central, and mixed events would allow for individuals who do not tolerate contact sensors (e.g. young children and individuals with intellectual disabilities) to be assessed for sleep apnea. Sleep apnea is underdiagnosed and undertreated in these populations due to poor patient ability to tolerate current PSG or HSAT~\cite{parakh_sleep_2021}. A contactless method of evaluation for sleep apnea may allow for repeat studies to be performed more easily to assess how well an intervention (e.g., sleeping on one’s side, using an oral appliance, or undergoing a surgical procedure) changes one's sleep apnea. This would allow patients to try different management techniques and see what works best for them. Furthermore, information could be transmitted wirelessly if interpretation by a technician is required or automatically through the use of an analysis app. Therefore, finding alternatives to conventional sleep monitoring that are both non-contact and can be conducted in a patient's home can greatly improve patient care and speed of diagnosis. Inspired by the use of multimodal camera+radar setup for remote vital sensing~\cite{vilesov2022blending}, we investigate using thermography-based nasal airflow measurements as a replacement for the contact nasal airflow sensors and radar-based respiratory effort measurements as a replacement for chest motion sensors. We present the following contributions of this work:
\begin{enumerate}
    \item A comparison of radar and thermal modalities for remote sleep monitoring detection of breathing and sleep apnea. 
    \item A non-contact multimodal thermal and radar stack for detection and clinically relevant classification of apnea.
    \item A dataset composed of 10 sleeping patients with hardware-synchronized thermal videos, frequency-modulated continuous wave (FMCW) radar data, and ground truth waveforms and annotations by a certified sleep technician at a sleep center. In addition, we open-source our data-collection framework, code base, and circuit schematics for collecting hardware-synchronized radar and thermal data.
\end{enumerate}

\section{Related Works}

\subsection{Thermography for Airflow Estimation}
\par Thermal imaging has been explored for many medical applications~\cite{zhao2023making, mambou2018breast, ammer1996diagnosis, godoy2017detection}. In clinical PSG, a thermistor measures nasal airflow by detecting the thermal fluctuations near nostrils due to breathing~\cite{troester_aasm_2023}. These fluctuations are also visible in thermal videos of a patient’s face~\cite{barbosa_pereira_estimation_2017, chan_estimation_2020,yang_graph-based_2022}. Mozafari et al. decompose the videos into rank-1 tensors and perform a spectral analysis on their power spectral densities to estimate the signal~\cite{mozafari_respiration_2022}. Szankin et al. uses a slow-fast convolutional neural network~\cite{feichtenhofer_slowfast_2019} as a regressor to estimate the airflow signal~\cite{szankin_thermal_2023}. Methods for estimating airflow from thermal videos have also been validated on newborns~\cite{pereira_noncontact_2019} and preterm infants~\cite{pereira_estimation_2017}. In addition to the previously discussed unimodal methods, several multimodal methods have been proposed that augment low-resolution thermal videos with RGB~\cite{chen_collaborative_2020, khan_joint_2024} and depth~\cite{prochazka_breathing_2017} cues for lower-cost hardware deployment. 

\begin{figure}[t]
    \centering
    \includegraphics[width=0.45\textwidth]{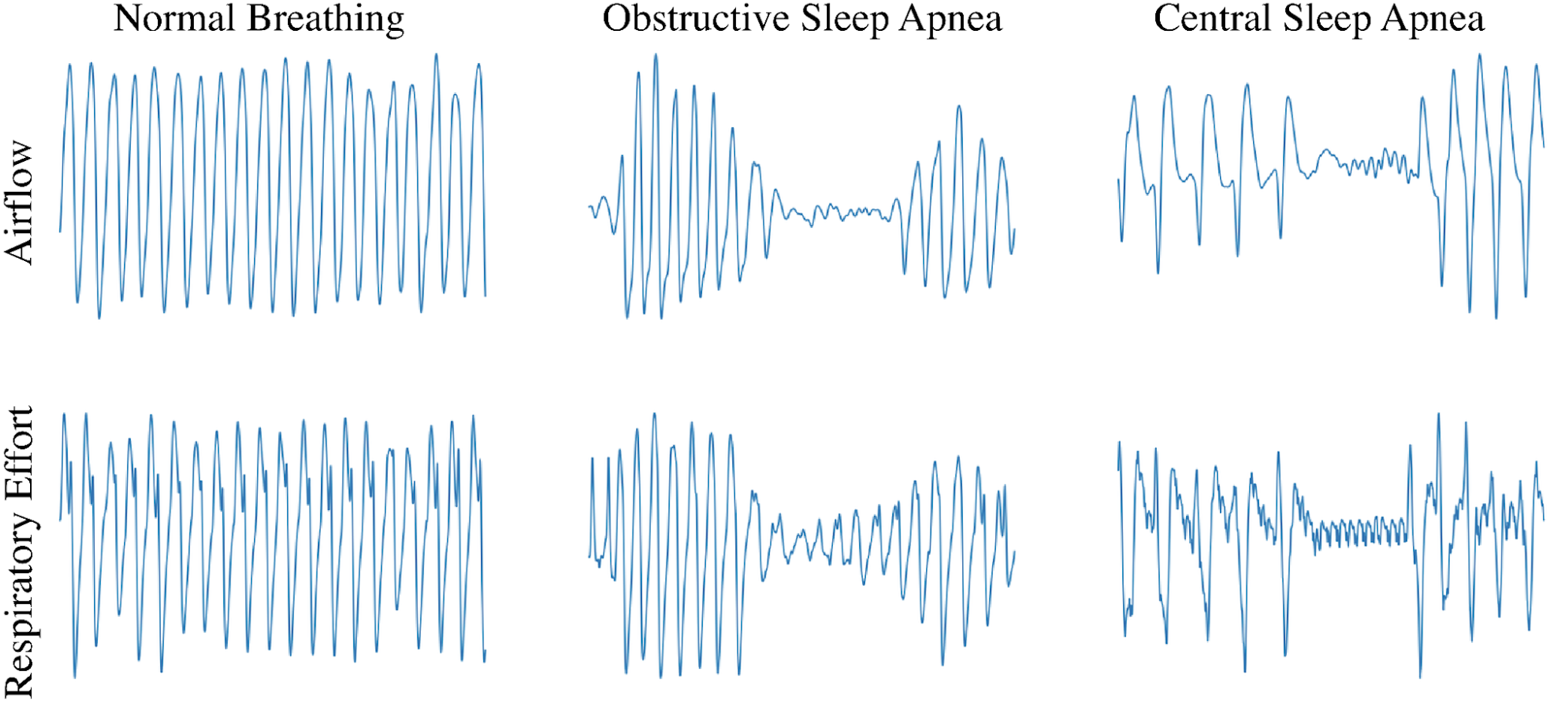}
    \caption{\textbf{Airflow and respiratory effort waveforms measured with sleep lab sensors depicting normal breathing, obstructive apnea, and central apnea.} During the onset of CSA, we notice anomalies in both the airflow and the respiratory effort waveforms; these anomalies manifest as attenuating factors that lower the amplitude of the two waveforms. Unlike CSA, the occurrence of OSA can be detected only through the airflow, which experiences a similar reduction in amplitude as the previous case. In comparison, the respiratory effort does not experience the same degree of attenuation for OSA.}
    \label{fig:example_gt_waveforms}
\end{figure}

\subsection{Visual and Wireless Sensing for Respiratory Effort}
\par Respiratory effort can be defined as the muscle movements of the chest that drive respiration~\cite{de_vries_assessing_2018}, visible to the human eye as the expansion of the chest, abdomen, and neck as the lungs fill with air. Clinically, they are detected using esophageal manometry or inductance plethysmography~\cite{troester_aasm_2023}. Several video-based algorithmic~\cite{braun_contactless_2018, prathosh_estimation_2017, gwak_motion-_2022, janssen_video-based_2015, 10.1145/2750858.2804280}, and data-driven approaches ~\cite{cheng_motion-robust_2023, alinovi_spatio-temporal_2015, alinovi_respiratory_2016} have been proposed to extract the respiratory effort signal. In addition to visual methods, thoracic and abdominal vibrations can be measured using wireless sensors, such as impulse-radio radars~\cite{khan_detailed_2017, zheng_more-fi_2021, husaini_non-contact_2022}, Doppler radars~\cite{zakrzewski_comparison_2012, lee_noncontact_2015, li_non-contact_2016, gu_assessment_2015}, and FMCW radars ~\cite{lv_non-contact_2021, purnomo_non-contact_2021, purnomo_non-contact_2022}. These approaches include both algorithmic ~\cite{lv_non-contact_2021, purnomo_non-contact_2021, li_non-contact_2016, gu_assessment_2015, zakrzewski_comparison_2012, khan_detailed_2017, guo_respiratory_2020} and data-driven \cite{zheng_more-fi_2021} solutions to estimate the respiratory effort. 

\begin{figure*}[t]
    \centering
    \includegraphics[width=1\textwidth]{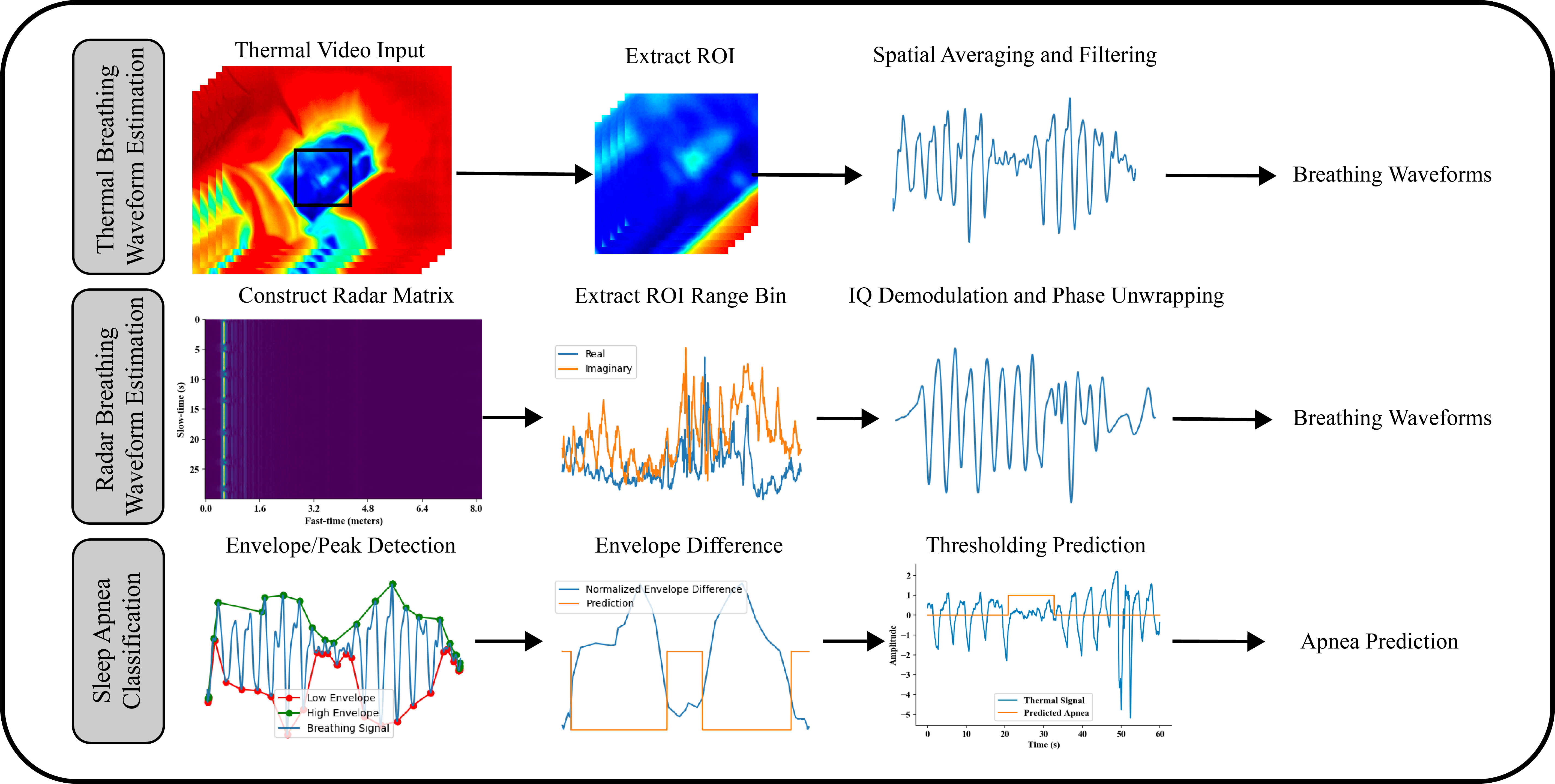}
    \caption{\textbf{Proposed pipeline for breathing detection from the radar and thermal modalities, followed by subsequent apnea detection.} First, we crop the frames acquired from the thermal camera (top) to focus only on the region near the nose. Then, we perform a global spatial averaging operation to collapse the video into a single time-series sequence, which, when filtered, gives us the breathing waveforms. For the radar (middle), we perform a standard Range-Doppler analysis to identify the approximate location of the patient, i.e., within a window of range bins. Once identified, we take an SNR-based weighted average of the extracted range bins to obtain breathing waveforms, which can then be filtered to increase waveform quality. Finally, we employ an envelope detection algorithm (bottom) to extract the upper and lower envelope, which can then be analyzed to detect anomalous regions on the waveform, i.e., regions with apnea.}
    \label{fig:pipeline}
\end{figure*}

\subsection{Automatic Apnea Detection}
\par Several methods have been developed to detect sleep apneas from ground truth breathing data using wavelet features ~\cite{fontenla-romero_new_2005,emin_tagluk_classification_2010, hassan_computer-aided_2016, avci_sleep_2015}, and neural networks~(e.g., MLPs, ANNs, CNNs, and LSTMs)~\cite{fontenla-romero_new_2005, zhang_automatic_2021, emin_tagluk_classification_2010, yue_deep_2021} of many architectures and envelope detection ~\cite{uddin_novel_2021}. Methods have also been developed to detect sleep apneas from features of heart rate waveforms \cite{bailon_ptt_2010, nguyen_online_2014, gutta_cardio_2018, gutta_model_2016}.

\par Several methods have been developed using nasal airflow information from infrared thermography to detect apneas in conjunction with either signal processing ~\cite{murthy_remote_2007, murthy_thermal_2009, fei_thermal_2009} or deep learning ~\cite{chung_camera_2023, hu_combination_2018}. An et al. propose a method to detect sleep apneas using nasal airflow information from infrared optical gas imaging \cite{an_non-contact_2022}. Parallel work has also been conducted to detect apnea events from acoustic recordings ~\cite{narayan_noncontact_2019, nakano_monitoring_nodate, xue_non-contact_2020, dafna_automatic_2013, nandakumar_contactless_2015, rosenwein_detection_2014, rosenwein_breath-by-breath_2015}. Ambient recordings of the patient have been analyzed to identify breathing signals and isolate periods where the breathing stops or is obstructed. While they can be effective for apnea detection, these methods do not directly measure the same clinically relevant physiological signals as the AASM recommended contact sensors \cite{troester_aasm_2023}. 

\par Kang et al. provide a method for detecting apnea events using a respiratory effort signal measured with an impulse-radio radar only~\cite{kang_non-contact_2020}. A high-quality signal is extracted by performing a range-doppler analysis, followed by a Kalman filtering operation~\cite{khan_detailed_2017}. Binary classifiers are then trained to predict apneas based only on this estimated respiratory effort signal ~\cite{kang_non-contact_2020}. Akbarian et al. provide a method for detecting apnea events from near-infrared videos of patients by computing the optical flow between frames of the videos and using a convolutional neural network to classify 10-second durations of the optical flow between apneic and non-apneic breathing with technician-annotated data as supervision~\cite{akbarian_noncontact_2021}. Carter et al provide a method for detecting apnea events using both respiratory and photoplethysmograph waveforms extracted from near-infrared videos \cite{deeplearningena-2023/6}

\par Non-contact methods have been developed to remotely measure SpO2. The most common method to achieve remote SpO2 extraction is through two NIR cameras with different wavelength sensitivities. Typically, these methods first detect a remote-photoplethysmography signal~\cite{vilesov2022blending, wang_remote_2020, chari2020diverse, paruchuri2023motion, liu2023efficientphys, chen2018deepphys, chari2024implicit}, followed by application of the ratio-of-ratios method~\cite{chan2013pulse, nitzan2014pulse}. This general framework has been applied in several instances~\cite{mathew2023remote, vogels2018fully, tian2022multi} with various algorithmic innovations to extract a better remote photoplethysmography signal. Other work also detects SpO2 using spectroscopic methods with a multi-aperture camera \cite{ramella2007spectroscopic, ramella2008measurement, ibrahim_assessment_2015}. While SpO2 is an important vital sign for sleep monitoring, we determined that we would require custom ground truth SpO2 measurement hardware to implement a noncontact method. This would have been incompatible with our data collection procedures as it would require our hardware to contact the patients.

\section{Background}
We provide a brief background of the primary mechanisms of the thermal and radar modalities relevant to understanding this work in sections \cref{ssec:thermal_bg} and \cref{ssec:radar_bg}. 

\subsection{Thermal Imaging}
\label{ssec:thermal_bg}

We overview the thermal camera image signal processing and underlying physics used to estimate changes in the temperature of the human body. All objects with a temperature higher than absolute zero ($-273.15^\circ C$) emit electromagnetic radiation that scales with the temperature of the body. The relationship is described by Planck's law. For a given wavelength $\lambda$, an object with emissivity $\varepsilon$ and temperature $T$ has spectral radiant exitance $I(\lambda, T)$~\cite{vollmer2020infrared}:
\begin{equation}
    I(\lambda, T) = \frac{2\pi \varepsilon hc^2}{\lambda^5} \frac{1}{e^{hc/(\lambda kT)} - 1} \;\;\; W m^{-2},
    \label{eq:planck}
\end{equation}
where $h=6.63\cdot 10^{-34} J\, s$ (Planck's constant), $k=1.38\cdot 10^{-23} J \, K^{-1}$ (Boltzmann constant) and $c=3\cdot 10^8 m \, s^{-1}$. However, since cameras integrate incoming radiance over a range of wavelengths, \cref{eq:planck} can be rewritten to obtain the total radiant flux emitted by a surface as:
\begin{align}
   I(T) & =\int_\lambda M(\lambda, T) \, d\lambda = \varepsilon \sigma T^4,
   \label{eq:stefan-boltzmann}
\end{align}
also known as the \emph{Stefan-Boltzmann law}, where $\sigma=5.67\cdot 10^{-8} W \, m^{-2} \, K^{-4}$ is the \emph{Stefan-Boltzmann constant}. The thermal radiation is proportional to the fourth power of temperature scaled by emissivity. Estimation of temperature can proceed by taking the fourth root of~\cref{eq:stefan-boltzmann} as.
\begin{equation}
    T = \sqrt[4]{ \frac{I(T)}{\varepsilon \sigma}}.
    \label{eq:singleobjectfourthroot}
\end{equation}

\begin{figure}[t]
    \centering
    \includegraphics[width=0.45\textwidth]{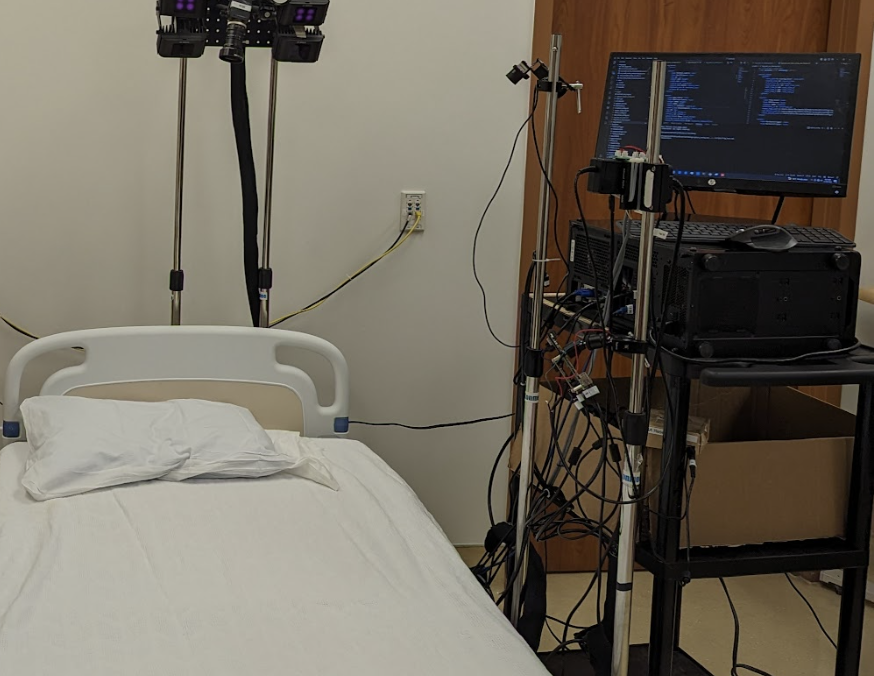}
    \caption{\textbf{Our experimental hardware setup consisting of a thermal camera, radar, and data-processing auxiliaries located in a sleep lab.} In this particular setting, we have placed the thermal camera and the radar to the right of the bed. Our setup also includes a microcontroller that is connected to the existing in-lab PSG hardware. This microcontroller sends pulse trains that can be used to synchronize the ground truth annotations with the captured recordings.}
    \label{fig:hardware_setup}
\end{figure}

Thermal cameras are excellent tools to measure changes in temperature in an environment, regardless of whether the camera is radiometric or not. For humans, most of this power is located in the infrared region which constrains thermal cameras to sense in the 8-14 $\mu m$ region. Several prior works ~\cite{mozafari_respiration_2022, yang_graph-based_2022, szankin_thermal_2023, barbosa_pereira_estimation_2017, chan_estimation_2020} have shown that thermal cameras can be used to detect the respiratory waveform due to the rhythmic change in temperature around the facial region due to breathing. For further details about thermal imaging characteristics and methodology, we defer the reader to Vollmer et al. ~\cite{vollmer2020infrared}.

\subsection{Radar}
\label{ssec:radar_bg}
Frequency modulated continuous wave (FMCW) radar emits and receives (reflected) chirps, that are linearly frequency modulated electromagnetic (EM) waves. The unique framework of FMCW radars modulates the transmitted chirp with the received chirp to produce a signal that allows for rough estimation of absolute distance on a centimeter resolution by observing its frequency as well as measuring fine-grained changes in distance at micrometer resolutions 
 by observing changes in the phase of the chirp.  The transmitted and received signal, $s(t)$ and $u(t)$ can be modeled as: 
\begin{equation}
    s(t) = A_s cos(2\pi f_c t + \pi k t^2), 0 < t < T_{c}.
\end{equation}
\begin{equation}
    u(t) = A_u cos(2\pi f_c (t-t_d) + \pi k (t-t_d)^2), t_d < t < T_{c}. 
\end{equation}
where $k$ is the frequency slope (the rate of change of frequency of the chirp), $f_c$ is the starting frequency of the chirp, $T_c$ is the duration of the chirp transmission, and $t_d$ is the time delay between the start of transmission and arrival of the reflection. 
% The bandwidth of the signal is the difference between the maximum and minimum frequencies, given by: 
% \begin{equation}
%     B = f_{max}-f_{min} = (f_c+kT_c) - f_c = kT_c,
% \end{equation}
The time delay is proportional to the round trip distance, $t_d = \frac{2R}{c}$, where $R$ and $c$ are the range of the object and speed of light respectively.

The radar modulates the received chirp with the still transmitting signal. The resulting signal is proportional to $s(t) \cdot u(t)$ and contains 2 components: a beat signal component with a frequency equal to the frequency difference of $s(t)$ and $r(t)$, $\Delta f = k t_d$, and a high frequency component situated near $4\pi f_c$. The higher frequency component is filtered out and thus generating $m(t)$. For brevity, the following equations are represented with just the in-phase component. The signal $m(t)$, can then be written as:
\begin{equation}
    m(t) \propto cos(2\pi f_c t_d + 2 \pi (k t_d) t + \pi k t_{d}^2), t_d < t < T_{c}.
\label{eqn:rf_if}
\end{equation}
Equation \ref{eqn:rf_if} can be rewritten into a more succinct form:
\begin{equation}
    m(t) \propto cos \left(\omega t + \phi\right), t_d < t < T_{c},
\label{eqn:rf_if_final}
\end{equation}
\begin{equation}
    \omega = 4 \pi\frac{kR}{c},\quad \phi = 4\pi \frac{R}{\lambda}.
\label{eqn:rf_if_final}
\end{equation}
The signal's phase and frequency depend on the range, $R$. They can be extracted through a discrete Fourier transform (DFT) of the signal after being passed through an analog to digital converter (ADC). The frequency term, $\omega=2\pi \Delta f$, provides the range through the following relation $R = c\frac{\Delta f}{2k}$. The phase term $\phi$ is inversely proportional to the wavelength of the radar, $\lambda=\frac{c}{f_c}$. The range of an object can be parameterized as $R(t) = R_{o} + r(t)$, where $r(t)$ models changes due to vibrations. To extract a breathing rate, $r(t)$ needs to be sampled with multiple chirps, thus creating a range matrix that is depicted in \cref{fig:pipeline}. Typically, the frequency term cannot be used to extract the possibly sub-centimeter displacement of the chest since the frequency resolution is on the order of centimeters. Instead, we use the highly sensitive phase to determine the oscillations of $r(t)$~\cite{alizadeh2019remote}.

\section{Methods}
\par We begin by describing breathing waveform and respiratory rate (RR) extraction in \cref{ssec:breathing_estimation}, followed by apnea detection in \cref{ssec:apnea_detection} for both the radar and thermal modalities. An overview of the process is shown in \cref{fig:pipeline}. We conclude with a description of our proposed apnea classification method in \cref{ssec:classify}.
\begin{table}[t]
    \centering
    \caption{Performance of Breathing Rate Estimation.}
    \label{tab:breathing_table}
    \begin{tabular}{lccccc}
        \toprule
        \textbf{Method} & \textbf{MAE} & \textbf{RMSE} & \textbf{MAPE} \\
        \midrule
        Alizadeh~\cite{alizadeh2019remote} & 4.91 & 6.97 & 32.56\% \\
        \textbf{Our Radar} & 1.90 & 3.95 & 14.45\% \\
        \midrule
        Chan~\cite{chan_estimation_2020} & 3.17 & 5.21 & 20.58\% \\
        \textbf{Our Thermal Method} & \textbf{1.58} & \textbf{3.45} & \textbf{11.88\%}  \\
        \bottomrule
    \end{tabular}
\end{table}

\begin{figure*}[ht]
    \centering
    \includegraphics[width=1\textwidth]{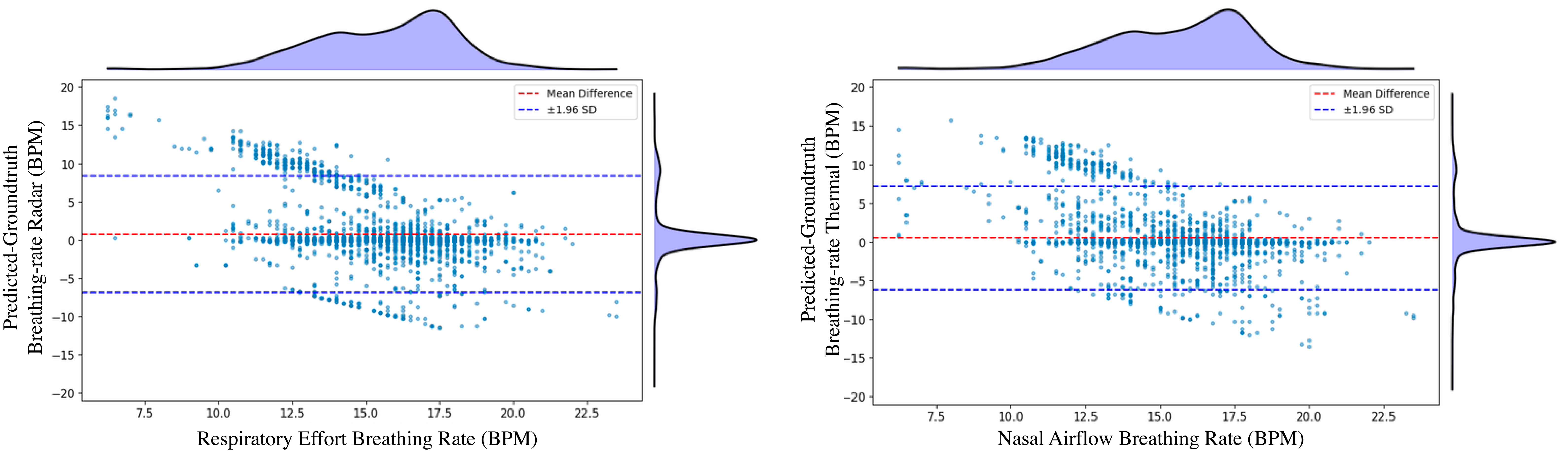}
    \caption{\textbf{Bland-Altman plot for breathing rates estimated by our radar (left) and thermal (right) modalities.}
}
    \label{fig:blandaltman}
\end{figure*}

\subsection{Breathing and Respiratory Rate (RR) Estimation}
\label{ssec:breathing_estimation}
\par While both thermal cameras and radars yield RR estimates, they do so by sensing slightly different physical phenomena. Thermal cameras monitor intensity changes, while radars track the instantaneous displacement of the chest (and/or the abdomen) to calculate the breathing rate.

\subsubsection{Pre-processing Thermal Recordings} The breathing rate information is primarily located in a small region below the nostrils caused by temperature changes during inhalation of room temperature air or exhalation of warm air from the lungs. Therefore, all videos of patients are manually cropped to a tight region around the nose as shown in \cref{fig:pipeline}.

\subsubsection{Airflow and RR from Thermography} A simple approach taken by prior works~\cite{chan_estimation_2020, barbosa_pereira_estimation_2017} collapse the video to obtain a 1D temporal signal, $x[t]$, by spatially averaging each frame. This is followed by filtering operations to limit frequencies to the accepted range of breathing rates. However, this approach does not translate perfectly to our setting because we do not have control over patient motion, which severely degrades the signal. Since motion and disturbances can be modeled as spikes or delta functions, we can compensate for them by averaging the derivative signal over a $N=25$ frame window following by a derivative operation to remove low frequency trends. We found that this operation dampens artifacts from motion as well as compensates for any spurious calibrations that the thermal camera requires. The operation can be written as:
\begin{equation}
\label{eq:motion_comp}
    y[t] = \frac{1}{N}\sum_{i=-\lfloor N/2 \rfloor}^{\lfloor N/2 \rfloor} x[t+i]-x[t+i-1].
\end{equation}

\subsubsection{Respiratory Effort and RR from Radar Sensors} The initial processing of FMCW radar data is described in \cref{ssec:radar_bg}. Here, we describe additional processing steps that are required for the detection of respiratory effort in a sleeping patient. Once a range matrix is constructed, the first step is to find the primary range bin that a person is located in. This is usually chosen as the range bin with the maximum power~\cite{alizadeh2019remote}. However, this assumes that the patient is the main object in view, which is not always the case in a sleep monitoring setting where the radar is placed on the side of the patient. We improve upon this by taking a window of range bins, $M$, around the maximum power range bin and choosing the range bin with the maximum SNR. We calculate the SNR of of the $i$th unwrapped range bin, $x_i[t]$ as:
\begin{equation}
    \alpha_i = \frac{\sum_{f \in F_{signal}}|X_i[f]|^2}{\sum_{f \in F_{noise}}|X_i[f]|^2},
\end{equation}
where $X_i[f]$ is the DFT representation of $x_i[t]$, $F_{signal}$ is a small set of frequency bins centered around the frequency bin that contains the most power in the range of $0.1-0.5$ Hz, and $F_{noise}$ contains the remaining frequency bins in the breathing frequencies. The final breathing signal, $y[t]$ is:
\begin{equation}
    y[t] = \alpha_{i^*}\cdot x_{i^*}[t].
\end{equation}
where $i^* = argmax_i \ \alpha_i$. We additionally process this signal with \cref{eq:motion_comp} to also help with motion and phase unwrapping artifacts. 

\begin{table*}[t]
    \centering
    \caption{Performance of Sleep Apnea Detection}
    \label{tab:performance_comparison}
    \begin{tabular}{lccccc}
        \toprule
        \textbf{Method} & \textbf{Accuracy} & \textbf{Precision} & \textbf{Recall} & \textbf{F1 Score} & \textbf{Intra-Class Correlation} \\
        \midrule
        \textbf{Our Radar Method} & 0.88 & 0.12 & 0.84 & 0.21 & 0.14 \\
        \textbf{Our Radar Method (With Motion Detection)} & 0.83 & 0.13 & \textbf{0.86} & 0.22 &  0.13\\
        % Kang's Method & acc & pre & rec & f1 & icc \\
        \midrule
        \textbf{Our Thermal Method (With Motion Detection)} & 0.90 & 0.19 & 0.79 & 0.31 & 0.26 \\
        \textbf{Our Thermal Method} & \textbf{0.99} & \textbf{0.68} & 0.74 & \textbf{0.71} & \textbf{0.70} \\
        \bottomrule
    \end{tabular}
\end{table*}

\subsection{Sleep Apnea Detection}
\label{ssec:apnea_detection}
\par We employ an envelope detection algorithm for detecting sleep apnea events remotely using a thermal camera or a radar sensor. With conventional algorithms, such as the Hilbert Transform, limited to narrow-band fluctuations~\cite{ciolek2014automated}, we instead opt to detect critical points in the signal and process them for continuous predictions of the lower and upper envelopes of the signal. In addition to envelop detection, detecting motion is crucial for detection to filter false positives.

\subsubsection{Motion Detection} 
\par After the thermal video is captured, it is processed and cropped in real time to eliminate any artifacts related to the background of the video. In some instances, significant amounts of patient motion can hinder the data quality of the thermal camera due to the lack of visibility of ora-nasal airflow from the nose of the patient. To combat this, we designed a peak-based motion detection algorithm.

A common way patient motion manifests itself in the breathing signal is in the form of singular high-magnitude peaks that significantly alter the envelope of the thermal signal. In order to filter out these peaks, for each key point, we compute the average distance to its $K$ nearest neighbors and filter out signal chunks surrounding the key point whose distance metric is unusually high. That is, given an array of peaks, $s[p] \in \mathbb{R}^{P}$ with $P$ local minima or maxima, the filtered set of peaks is given by:
\begin{equation}
    \left\{s[i] \ | \ \frac{\sum^{k = i - 1 + \frac{K}{2}}_{k = i - \frac{K}{2}} |s[k+1] - s[k]|}{K} > \beta=2.5 \right\}
\end{equation}

\subsubsection{Envelope Detection}
\par After filtering for motion, we extract the breathing signal as described in \cref{ssec:breathing_estimation}. To find the envelope of the breathing signal, $s$, we need to determine its keypoints. We reconstruct the lower envelope with the minima points and the upper envelope with maxima points. The noisy nature of the signal necessitates a denoising operation, both pre and post-keypoint detection. Once we filter out unwanted key points, a continuous version of the lower and upper envelopes of the signal is constructed via linear interpolation.  We then use the envelope difference normalized by its mean for apnea detection.

\par Several data-driven methods~\cite{yuzer_different_2020, fontenla-romero_new_2005, van_steenkiste_automated_2019, zhang_automatic_2021,yue_deep_2021,avci_sleep_2015} also exist for detecting sleep apneas from contact-based respiratory signals. However, due to the limited size of our non-contact dataset, we cannot replicate machine-learning driven algorithms.

\subsection{Sleep Apnea Classification}
\label{ssec:classify}
Differentiating between OSA and CSA is difficult with access to only one modality. However, the reader may notice in \cref{fig:example_gt_waveforms} that during a CSA, both the respiratory effort and nasal airflow signals decrease in amplitude, while for OSA, only the nasal airflow decreases in amplitude. We propose leveraging this observation to perform sleep apnea classification using both the thermal and radar modalities, where the remote sensors replace the nasal airflow and respiratory effort sensors, respectively. Classification then reduces to simple boolean algebra. Given the apnea predictions from radar, $y_{Radar}(t)$, and thermal, $y_{Thermal}(t)$, (where $y(t)=1$ and $y(t)=0$ denote apnea present and no apnea present, respectively), then we can formula CSA and OSA classification as: 
\begin{equation}
\label{eq:CSA}
    y_{CSA}(t) = y_{Radar}(t)\cdot y_{Thermal}(t)
\end{equation}
\begin{equation}
\label{eq:OSA}
    y_{OSA}(t) = y_{Thermal}(t)\cdot(1-y_{Radar}(t))
\end{equation}

\section{Experiments and Results}
\subsection{Hardware Setup}
\par As part of our clinical validation, six-hour recordings of patients participating in their PSG study were captured. Our hardware setup primarily consists of a thermal camera and radar placed in the periphery of the bed, visualized in \cref{fig:hardware_setup}. This is in addition to the existing ground truthing equipment used in PSG studies. 

\par A radiometrically calibrated Teledyne FLIR Boson with a $512 \times 640$ resolution was positioned to the side of the bed and aimed at the face. This choice of placement was necessitated by the constraints of existing PSG procedures. As a result, the patient’s nose was not always visible; for example, if they turned to the side facing opposite the camera, their face would be obscured. Beyond these experimental constraints, as a product set, we conjecture that a ceiling-mounted thermal camera could alleviate concerns about occlusions. 

\begin{figure*}[t]
    \centering
    \includegraphics[width=0.95\textwidth]{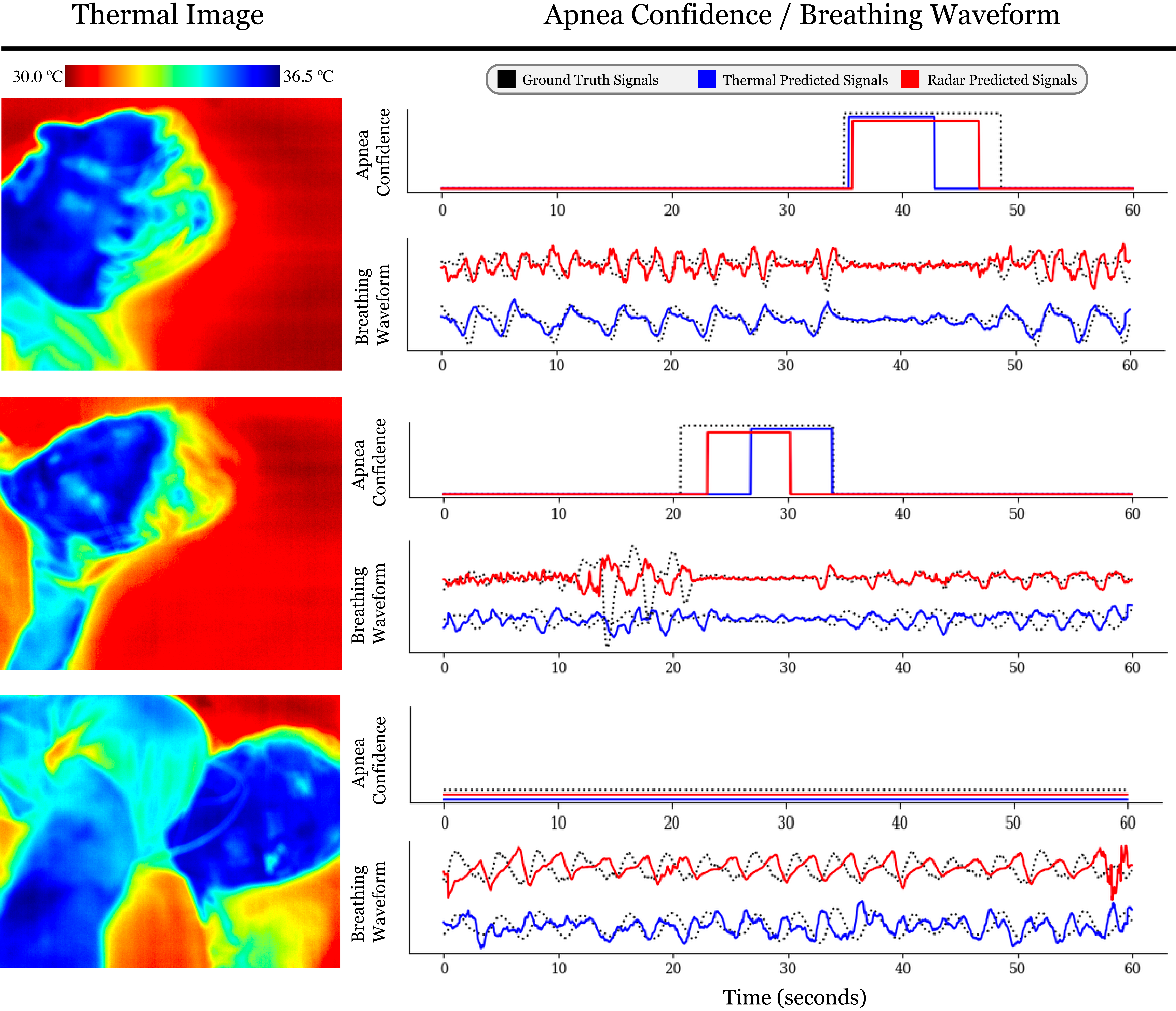}
    \caption{\textbf{Our apnea confidence scores and breathing waveforms are estimated from our thermal and radar recordings for several apnea events.} }
    \label{fig:big_results}
\end{figure*}

\begin{figure*}[t]
    \centering
    \includegraphics[width=0.9\textwidth]{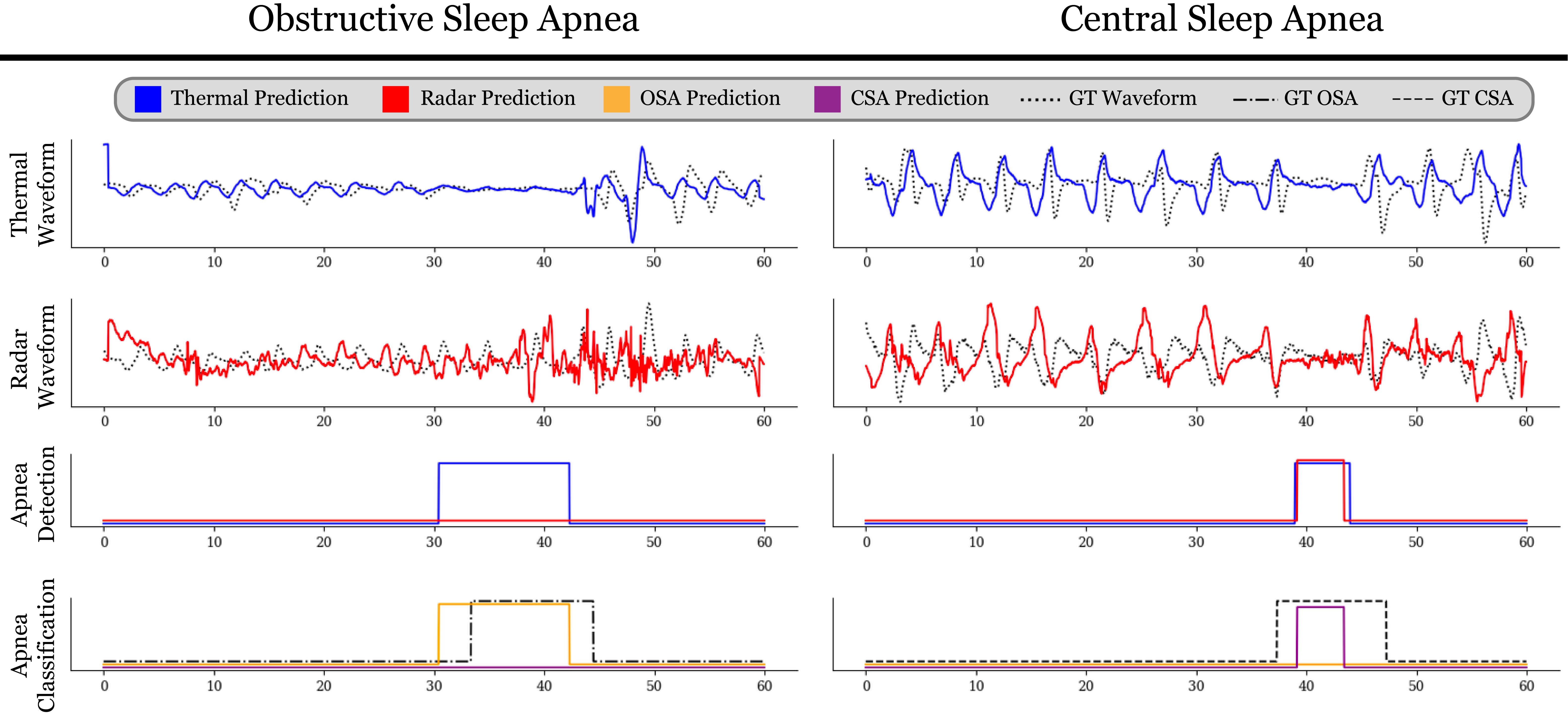}
    \caption{\textbf{A qualitative comparison between how obstructive and central sleep apnea appear to the radar and thermal modalities.} The thermal modality reduces in amplitude for both OSA and CSA, but the radar modality reduces in amplitude only for CSA. This distinction can be used to classify between OSA and CSA in a multimodal setup using radar and thermal. Apnea classification can then be performed by applying \cref{eq:CSA} and \cref{eq:OSA} with filtering of detections that are too short.}
    \label{fig:osa_csa_comparison}
\end{figure*}

\par We also place an AWR1443BOOST FMCW radar beside the thermal camera and in the periphery of the patient, next to the bed. Similar placement constraints exist for the radar; however, unlike the thermal camera, the radar is not adversely affected by the orientation of the patient. We balance attenuating factors from lateral position shifts and elevation offsets by operating the radar with all 3 transmitters and 4 receivers enabled. This allows us to perform beamforming post-acquisition and improve the SNR.

\par A synchronization signal was sent via an Arduino microcontroller to align the ground truth signals recorded by the external hardware used in contemporary PSG studies, as well as triggering the thermal and radar sensors at a 30 Hz rate. The alignment was performed on the vital sign recordings, as well as other ground truth labeling obtained from the PSG and technicians involved with the sleep study. A full-night or split-night PSG was recorded and annotated by a trained sleep technician in accordance with AASM guidelines ~\cite{troester_aasm_2023}. 

\subsection{Dataset Details}
\par 10 patients with suspected OSA undergoing full-night or split-night PSG were enrolled in this study. Of these patients, 1 was diagnosed with mild OSA ($5 \leq AHI \leq 15$), 3 were diagnosed with moderate OSA ($15 \leq AHI \leq 30$), and 1 was diagnosed with severe OSA ($AHI \geq 30$). The patients provided their written informed consent in accordance with our Institutional Review Board permissions, and all methods were performed in compliance with relevant guidelines and regulations of the University of California, Los Angeles. This study was approved by the UCLA Institutional Review Board, IRB\#21-000018.

In our dataset, we discarded one patient's data due to synchronization error and another patient's data due to improper setup of the thermal camera. We also discarded periods of the recordings where the patients' noses were not in the frame of the thermal video. The entire dataset contains 20 hours and 25 minutes of synchronized thermal videos, radar recordings, and ground truth PSG recordings. Of the patients in the final dataset, two patients had sleep apnea events during the valid recording periods.

\par When calculating metrics in Table 1, we over-sampled 1-minute-long chunks of the data twenty times for every 5 minutes of data, resulting in an over-sampled dataset that is 59 hours 40 minutes long. Using our motion detection algorithm, we classified that 34 hours and 42 minutes of data in the over-sampled dataset do not contain significant levels of motion. 

\subsection{Results}
\paragraph{Breathing Estimation Evaluation} We present qualitative in \cref{tab:breathing_table} as well as in a Bland-Altman plot in \cref{fig:blandaltman}. We also show qualitative results of the estimated breathing waveforms in \cref{fig:big_results}. We find that the thermal modality outperforms the radar in breathing rate estimation. We hypothesize that this is due to reflections from the 77 GHz radar being specular. This can lead to a reduced signal when a patient's chest is not perpendicular to the radar's optical axis. We also find that our radar smoothing and SNR weighting scheme, \cref{ssec:breathing_estimation}, improves upon prior breathing estimation methods~\cite{alizadeh2019remote}.

\paragraph{Sleep Apnea Detection} We demonstrate the qualitative results in \cref{fig:big_results}. Overall, from \cref{tab:performance_comparison}, we found that thermal imaging provides more robust sensing of apneas than the radar when accounting for motion. While the radar was able to achieve the highest recall, it suffered from low precision. We believe that this is due to the sensitivity of the radar wave's phase to movement. Even small movements can cause changes in the amplitude of the signal, resulting in false positives. Using motion detection and handling is advantageous as it allows us to filter out parts of the signals that have anomalies due to patient motion. The motion handling algorithm removes adverse distribution shifts to the distribution of local maxima and minima that make up the envelope of the signal, causing our algorithm to classify false-positive apnea events. Furthermore, our motion handling algorithm is not only limited to our apnea detection algorithm but can also be used to inform other apnea detection or breathing rate estimation algorithms about the presence of motion and allow for proper handling. However, motion detection comes with the tradeoff of possibly discarding sleep apnea samples. For the final results, we used a threshold of 0.4 for thermal and 0.5 for radar for the motion detection algorithm and a window size of 23 for the upper and lower envelopes.

Two patients  had apnea events in the valid recording period. The first participant had 20 ground truth apneas (1 OSA and 19 CSA): 18 of the apneas were predicted by the thermal camera data and 16 of the apneas were predicted by the radar data by our algorithm. The second participant had 7 ground truth apneas (1 OSA and 6 CSA): 2 of the apneas were predicted by the thermal camera data and 4 of the apneas were predicted by the radar data. predicted apneas on the thermal camera data.

\paragraph{Sleep Apnea Classification} We also demonstrate an application of using both radar and thermal modalities for apnea classification between OSA and CSA. In \cref{fig:osa_csa_comparison}, we can see an example of OSA and CSA from our dataset classification according to \cref{eq:CSA} and \cref{eq:OSA}. Due to the small number of OSA and CSA examples in our dataset, we can only show qualitative multimodal results of OSA and CSA classification.  

\section{Limitations}
\par Due to our radar and thermal setup, we are not able to incorporate blood oxygenation estimation techniques, although the AASM guidelines recommend that a blood oxygenation sensor be included in both PSG and HSAT. Therefore, since blood oxygenation is a criterion for hypopnea classification, hypopnea detection using completely non-contact methods is not possible in alignment with AASM criteria. However, it may be possible to use relaxed criteria to score hypopneas, omitting the blood oxygenation information. Furthermore, recent work has suggested that oxygen saturation can be remotely acquired in limited settings~\cite{vogels2018fully, sun2024ccspo, hu_combination_2018}. We hope that in future work, we can incorporate these methods to enable full non-contact monitoring. Additionally, as previously stated in \cref{ssec:breathing_estimation}, extreme sleeping positions can obscure the nose from the thermal camera, the modality found to be crucial for apnea detection. While remote sensing has the potential to benefit patients, it is still a new technology that warrants further studies to understand generalizability and fairness~\cite{kadambi2021achieving} of the technology to diverse patient populations.  

\section{Conclusion}
We propose a novel method of contactless nasal airflow and respiratory effort estimation of apnea to function as a real-time drop-in replacement for existing contact sensors. We verify the performance of these methods using data collected from patients with suspected sleep apnea. We demonstrate that these methods can be used to detect sleep apnea events and even potentially distinguish between some obstructive and central apnea events. These methods could lead to the development of a portable, repeatable, non-contact diagnostic tool for populations underdiagnosed with sleep disorders due to their inability to access or tolerate current PSG and HSAT diagnostics.

\appendices


\section*{Acknowledgment}
The authors would like to thank UCLA Sleep Disorders Center and its staff, especially Julie Toomey and Weiguang Zhong; undergraduate students Rishabh Sharma, Rui Ma, Jianchong Ma, and Julia Craciun for their assistance in developing parts of the preprocessing code; Junaid Ahmad for his work designing protective circuits for data collection; medical student Clare Moffatt for help with data collection; and Dr. Laleh Jalilian for valuable advice in collecting clinical data. Authors on this research were supported by a DARPA Young Faculty Award; Achuta Kadambi was supported by a National Science Foundation (NSF) CAREER award IIS-2046737, an Army Young Investigator Program Award, and a Defense Advanced Research Projects Agency (DARPA) Young Faculty Award. 


\ifCLASSOPTIONcaptionsoff
  \newpage
\fi



%\bibliographystyle{IEEEtran}


% \end{thebibliography}
\bibliographystyle{acm}
\bibliography{OSA}



%\vfill


\end{document}